\documentclass[letterpaper, 10 pt, conference]{ieeeconf}  % Comment this line out if you need a4paper

\IEEEoverridecommandlockouts                              % This command is only needed if 
\usepackage{amsmath,amssymb,amsfonts}
\usepackage{balance}
\usepackage{cite}
\usepackage{algorithmic}
\usepackage{graphicx}
\usepackage{textcomp}
\usepackage{xcolor}
\usepackage{csquotes}
\usepackage{siunitx}
\usepackage{subcaption}
\usepackage{gensymb}
\usepackage{url}
\usepackage{multirow}
\usepackage{colortbl}
\usepackage{wasysym}
\usepackage{booktabs}
\let\labelindent\relax
\usepackage{enumitem}

\makeatletter
\let\NAT@parse\undefined
\makeatother
\usepackage{hyperref}
\hypersetup{
    colorlinks,
    linkcolor=[HTML]{B75659},
    citecolor=[HTML]{B75659},
    urlcolor=[HTML]{6595BF}
}

\title{\LARGE \bf
Instrumentation for Better Demonstrations: A Case Study
}

\author{Remko Proesmans$^{1}$, Thomas Lips$^{1}$ and Francis wyffels$^{1}$% <-this % stops a space
\thanks{*This work was supported by the Research Foundation Flanders (FWO) under grant agreement no. 1S15925N and 1S56024N,  and by the euROBIN Project (EU grant number 101070596). }% <-this % stops a space
\thanks{$^{1}$Remko Proesmans, Thomas Lips and Francis wyffels are with the AI and Robotics Lab (IDLab-AIRO), Ghent University---imec, Ghent, Belgium
        {\tt\small remko.proesmans@ugent.be}}%
}

\def\BibTeX{{\rm B\kern-.05em{\sc i\kern-.025em b}\kern-.08em
    T\kern-.1667em\lower.7ex\hbox{E}\kern-.125emX}}
\begin{document}
\bstctlcite{IEEEexample:BSTcontrol}

\maketitle
\thispagestyle{empty}
\pagestyle{empty}

%%%%%%%%%%%%%%%%%%%%%%%%%%%%%%%%%%%%%%%%%%%%%%%%%%%%%%%%%%%%%%%%%%%%%%%%%%%%%%%%
\begin{abstract}

Learning from demonstrations is a powerful paradigm for robot manipulation, but its effectiveness hinges on both the quantity and quality of the collected data. In this work, we present a case study of how instrumentation, i.e. integration of sensors, can improve the quality of demonstrations and automate data collection. We instrument a squeeze bottle with a pressure sensor to learn a liquid dispensing task, enabling automated data collection via a PI controller. Transformer-based policies trained on automated demonstrations outperform those trained on human data in 78\% of cases. Our findings indicate that instrumentation not only facilitates scalable data collection but also leads to better-performing policies, highlighting its potential in the pursuit of generalist robotic agents.

\end{abstract}

%%%%%%%%%%%%%%%%%%%%%%%%%%%%%%%%%%%%%%%%%%%%%%%%%%%%%%%%%%%%%%%%%%%%%%%%%%%%%%%%
\section{INTRODUCTION}
Although robots can surpass humans for specific tasks in carefully constructed environments, human-level performance is still beyond the reach of general-purpose robots in unstructured environments~\cite{hu2024generalpurposerobotsfoundationmodels}.
One way to approach human-level manipulation performance with robots is to make them imitate human behaviour through demonstrations. This has proven to be effective in teaching complex manipulation skills to robots \cite{act, chi2024diffusionpolicy, pi0, openvla}.
However, learning even a single task in a controlled environment can take 100 or more demonstrations~\cite{act, chi2024diffusionpolicy}.
For a robot to handle multiple complex tasks in unstructured environments, the data requirements grow tremendously. Large datasets for robot pre-training exist~\cite{openxembodiment, droidinthewilddata, bridgedatav2datasetrobot}, but current systems still require task-specific fine-tuning for complex tasks or diverse environments~\cite{pi0}.

In addition to larger datasets and algorithmic advances, we can attempt to improve the quality of the demonstrations to increase imitation learning capabilities.
Both~\cite{maram2022} and \cite{bilal2024} determined evaluation metrics for the quality of demonstration trajectories, and showed that better demonstrations can result in a better policy for the same amount of data.
Xu et al.~\cite{xu2024rldg} have shown that reinforcement learning (RL) agents generate high-quality trajectories through reward maximisation, making them better suited for fine-tuning generalist policies compared to human demonstrations.

A different approach was followed in~\cite{Junge2023} for the task of raspberry harvesting: instead of teleoperating the robot and recording trajectories as demonstrations, human demonstrators manipulated an instrumented, i.e. sensorised, strawberry phantom.
The robot behaviour was then tuned to match human behaviours as experienced by the strawberry.
Instrumentation was also used in~\cite{openai2019} for learning to solve a Rubik's cube with a single anthropomorphic robot hand. 
Specifically, the sensor data provided accurate state information of the Rubik's cube to shape the reward function.

\begin{figure}[t]
\centering
%\vspace{2.7mm}
\begin{subfigure}[t]{0.27\textwidth}
    \centering
    \includegraphics[height=3.3cm]{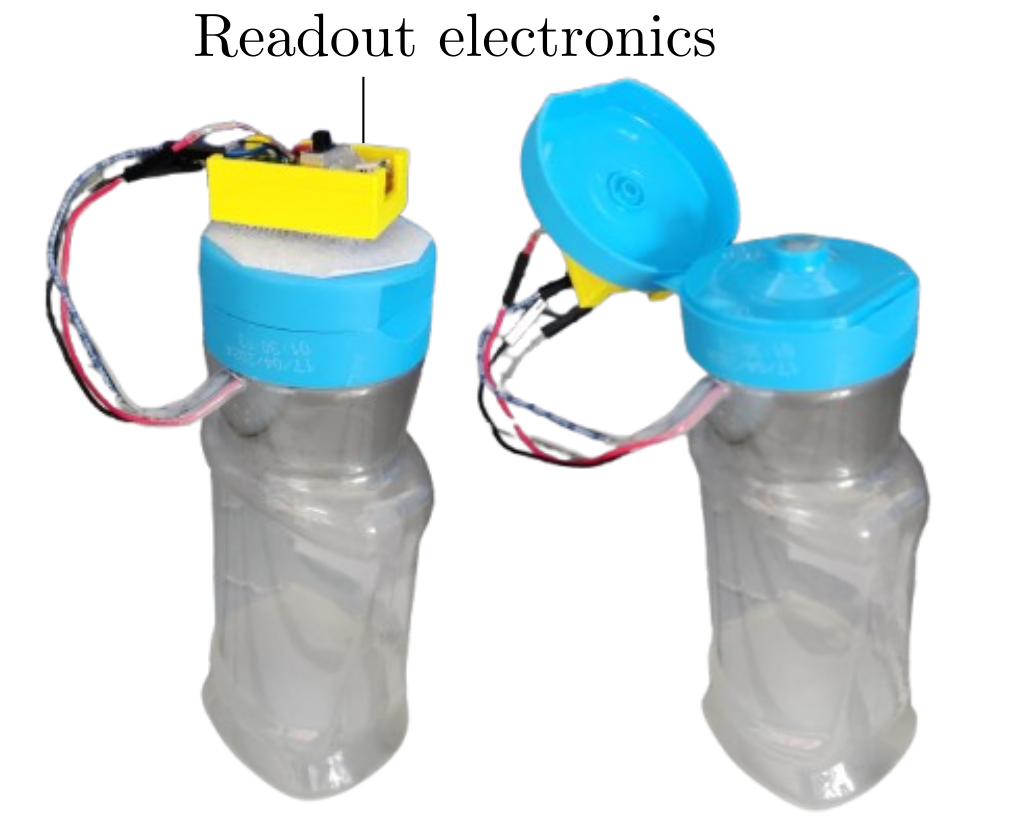} 
    \caption{Instrumented bottle used for autonomous data collection.}
    \label{fig:bottle_x2}
\end{subfigure}\hfill
\begin{subfigure}[t]{0.2\textwidth}
    \centering
    \includegraphics[height=3.3cm]{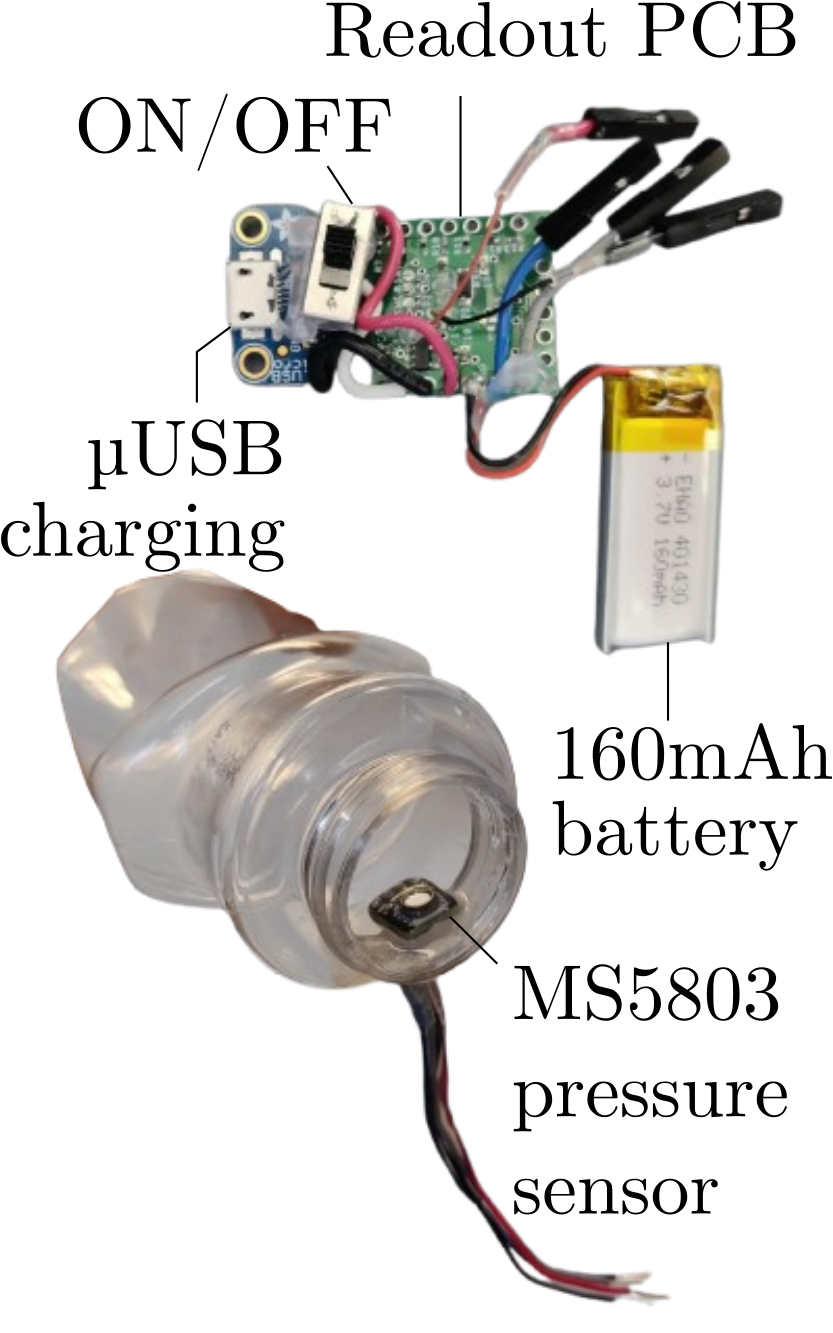}
    \caption{Electronics for the instrumented bottle.}
    \label{fig:bottle}
\end{subfigure}
\vspace{2mm}
\begin{subfigure}[t]{0.23\textwidth}
    \centering
    \includegraphics[height=3.3cm]{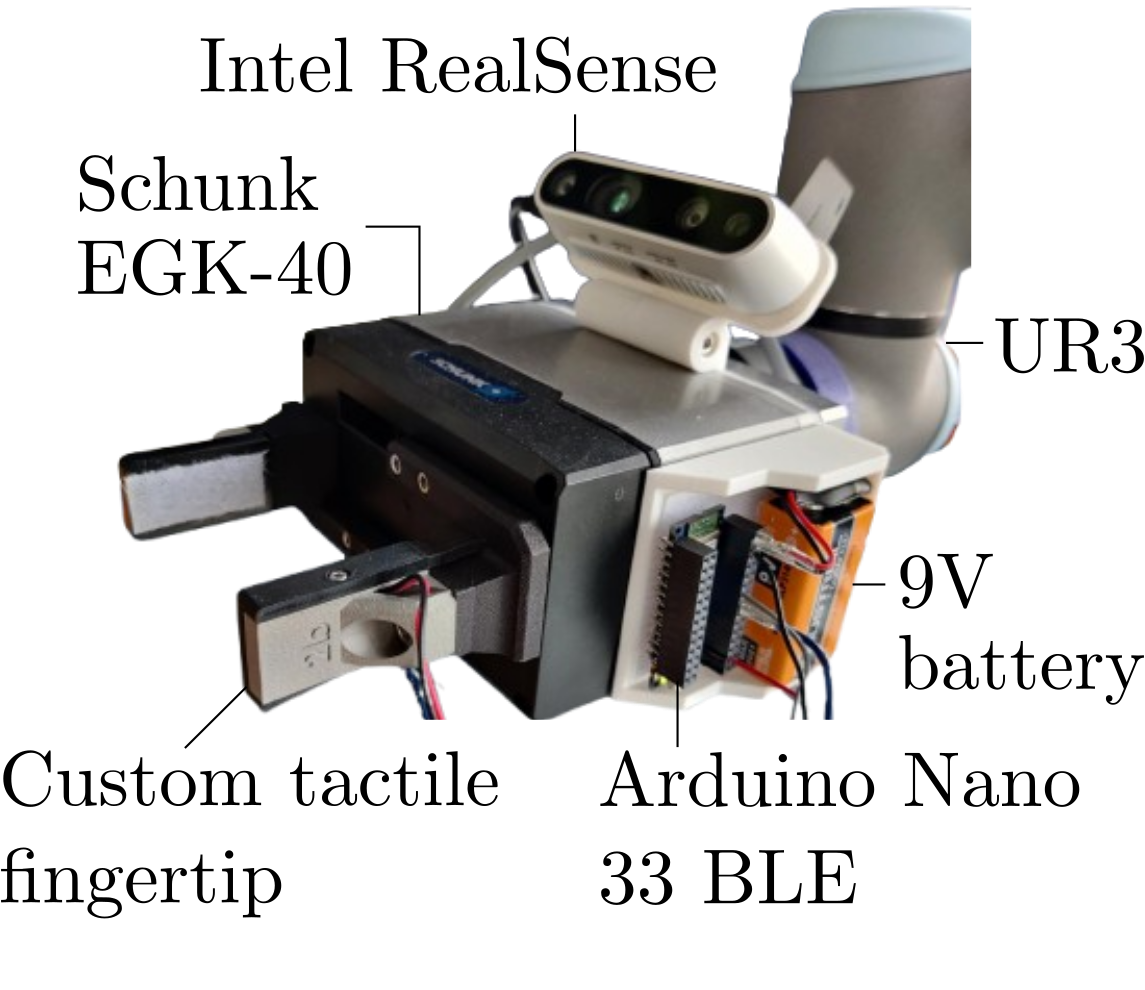} 
    \caption{Gripper mount.}
    \label{fig:gripper_closeup}
\end{subfigure}
\begin{subfigure}[t]{0.23\textwidth}
    \centering
    \includegraphics[height=3.3cm]{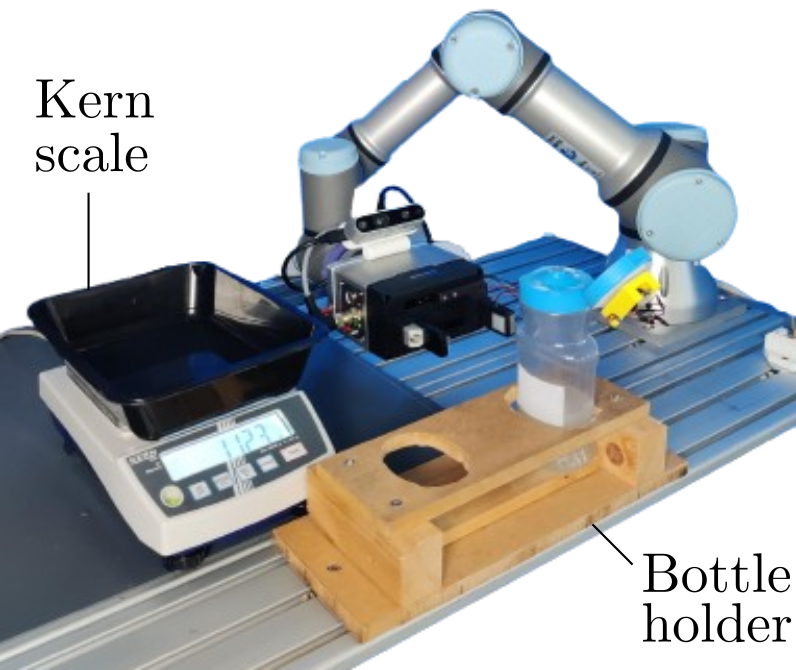}
    \caption{Robot setup.}
    \label{fig:setup1}
\end{subfigure}
\begin{subfigure}[t]{0.27\textwidth}
    \centering
    \includegraphics[height=3.3cm]{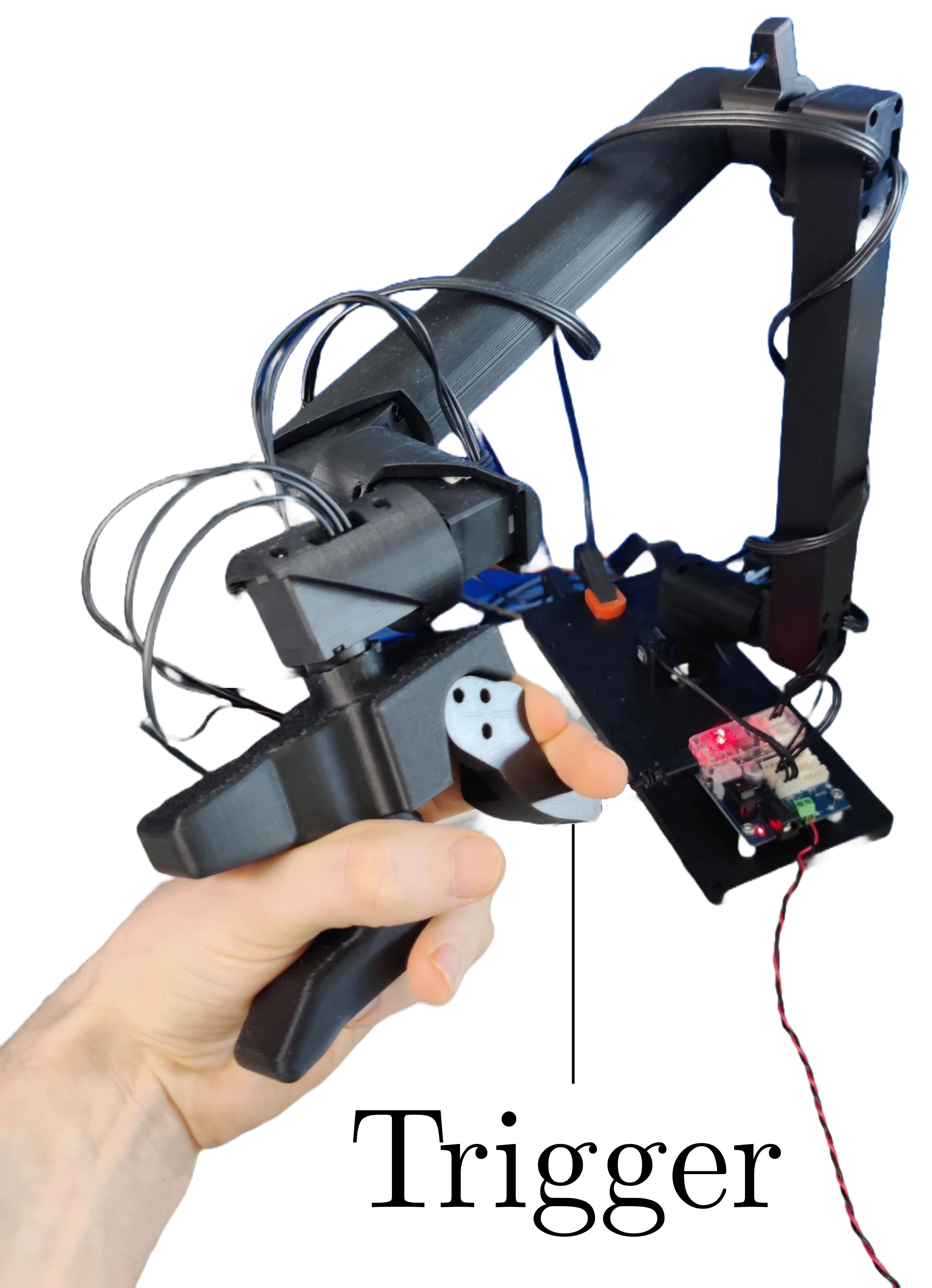}
    \caption{The Gello~\cite{gello} teleoperation arm.}
    \label{fig:teleop_closeup}
\end{subfigure}\hfill
\begin{subfigure}[t]{0.21\textwidth}
    \centering
    \includegraphics[height=3.3cm]{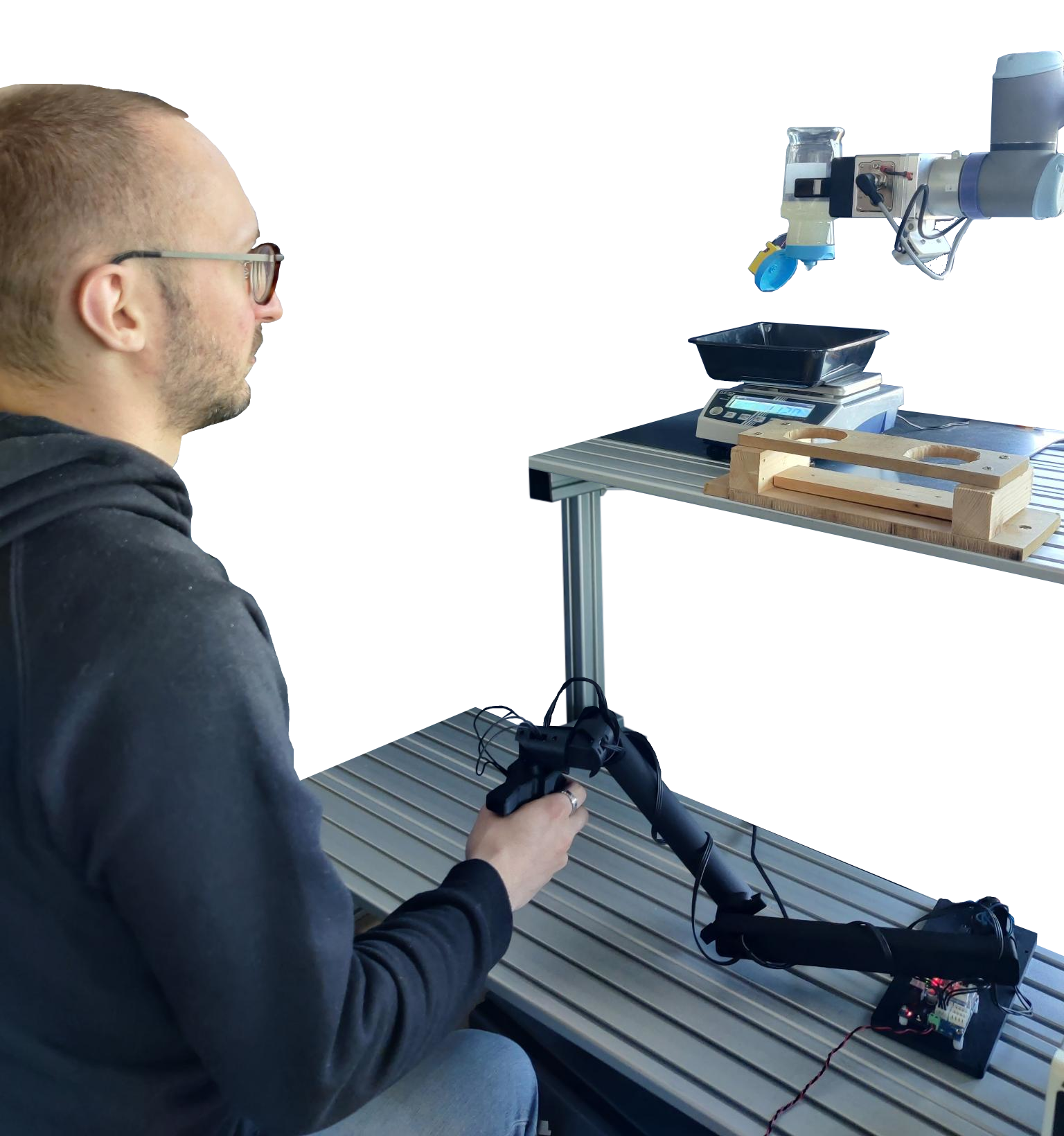} 
    \caption{Human operator.}
    \label{fig:teleop}
\end{subfigure}
\caption{Experimental setup for learning to squeeze a constant flow from a bottle based on demonstrations, collected either by a human or by an autonomous controller relying on instrumentation data.}
\label{fig:experimental_setup}
\end{figure}

The idea of using additional state information during the learning phase, and later deploying without this privileged information, using only sensory inputs that are available in the real-world setup, is also found in works that learn robot behaviours in simulation~\cite{lee2020,chen2023}. 
In~\cite{lee2020}, sim-to-real quadrupedal locomotion is achieved by first training a teacher policy that has access to privileged state information of the robot, allowing it to quickly achieve high performance. 
The teacher is then used to guide the learning of a student controller that only uses sensors available on the real robot.
Similarly in~\cite{chen2023}, the student-teacher paradigm is used to learn in-hand reorientation.
The teacher used the object orientation to quickly learn appropriate actions, after which the student was trained to match the teacher's outputs while relying on occluded visual observations of the objects.

Instrumentation like in \cite{Junge2023, openai2019} can be a way to obtain similar privileged information directly in the real world, potentially allowing for faster learning and/or more performant policies and avoiding the need for sim-to-real transfer, which already is a significant challenge on its own. 
We have previously applied it to learned cloth manipulation~\cite{verleysen2020} and state estimation~\cite{smarttextile}.
In this work, we present a case study on the use of instrumentation for demonstration learning, focusing on the task of squeezing a bottle to maintain a constant liquid flow.
We will show that (1) instrumentation can be used for automated collection of demonstrations by a \enquote{teacher} agent, (2) that these demonstrations are more qualitative than those collected by human teleoperators, and (3) that this quality difference improves the performance of a policy trained on those demonstrations, compared to training on human demonstrations.

%\begin{itemize}
%    \item We perform a case study of instrumentation for automated (and better) demonstrations
%\end{itemize}

\section{Materials \& Methods}

\subsection{Task Description}

With the aim of examining how instrumentation can be applied to demonstration learning, we chose the task of squeezing a constant flow of liquid from a bottle.
The flow rate is directly related to the pressure inside the bottle near the opening, so a constant flow can be achieved by integration of a pressure sensor and controlling its output.

%Other works that manipulated bottles: dosing squeeze bottle \cite{hellebrekers2020, huang2024}, pouring plastic bottle \cite{pozzi2022}, grasping \cite{li2021, kuppaswamy2020}
Plastic bottles are popular test objects in robotic manipulation~\cite{li2021, kuppaswamy2020, hellebrekers2020, pozzi2022, huang2024} because they are ubiquitous in human environments and deformable, which presents unique challenges for e.g. state observation. 
In~\cite{hellebrekers2020, huang2024} specifically, accurate dosing of liquids was tackled with tactile sensing.
Potential applications of this skill include culinary arts~\cite{huang2024} and the oiling or glueing of mechanical parts. 

\subsection{Hardware setup}

\subsubsection{Instrumented Bottle}
A commercial condiment squeeze bottle is instrumented by integration of a MS5803 waterproof pressure sensor, see Fig.~\ref{fig:bottle_x2}.
The supply and communication wires of the sensor are pulled through a small hole in the side of the bottle.
To make the sensor PCB as well as the cable hole waterproof, silicone (Silicone Addition Colourless 50 by Silicones and more) is applied all around it with a syringe.
The silicone is allowed to cure for 30\,minutes before application, so that it is viscous enough to not seep away immediately.
For readout, we repurposed a PCB from our open-source, modular, wireless Smart Textile~\cite{smarttextile}, featuring an nRF52840 microcontroller.
The PCB, along with a 160\,mAh LiPo battery, a micro USB charging interface, and an on/off button, is placed in a 3D printed holder and attached to the flip-top cap of the bottle with hook\nobreakdash-and\nobreakdash-loop tape.
The bottle is filled with a mixture of water and 2\,wt\% xanthan gum for thickening.
The bottle opening is a small hole without valve.

\subsubsection{Robot Setup}
A Schunk EGK\nobreakdash-40\nobreakdash-MB\nobreakdash-M\nobreakdash-B gripper is mounted to a UR3 robot, as seen in Fig.~\ref{fig:gripper_closeup}. 
One of the fingers is a standard Robotiq\,2F\nobreakdash-85 fingertip, the other is one of our custom tactile fingertips based on a magnetic transduction principle~\cite{magtouch}. 
The fingertip has a 2\nobreakdash-by\nobreakdash-2 grid of taxels with a 6\,mm pitch, each taxel outputs local 3D contact force estimation.
A custom 3D-printed adapter allows for attaching Robotiq-compatible fingertips to the Schunk.
In addition, an Intel RealSense 435 camera and an Arduino Nano 33 BLE (for readout of the tactile fingertip) are attached to the Schunk gripper via custom mounts.
The instrumented bottle is placed in a wooden holder before the robot grasps it.
When the bottle is grasped, it is turned over above a Kern~PCB2000\nobreakdash-1 scale.
The scale has a resolution of 0.1\,g.

\subsubsection{Teleoperation Setup}
The trigger of a Gello~\cite{gello} arm (Fig.~\ref{fig:teleop_closeup}) is configured to control the opening of the Schunk gripper.
A human operator (Fig.~\ref{fig:teleop}) squeezes the trigger while visually maintaining a constant flow.

\subsection{Data Collection Protocol}

Five different datasets are collected. 
For the first, different people manually squeeze the instrumented bottle in whatever way they deem fit, while attempting to maintain a constant flow by looking at the stream. 
This lasts 10 seconds, during which time the pressure in the bottle is recorded. 
Five people were each given a full bottle, and performed 10\,s trials until the bottle was empty.
This resulted in 34 trials.

The other four datasets involve the setup in Fig.~\ref{fig:experimental_setup}: a training dataset collected by a PI (PID with zero derivative gain) controller, a training dataset collected a human teleoperator, an evaluation dataset of rollouts from a policy $\Pi_{\text{PI}}$ trained on the PI data, and an evaluation dataset of rollouts from a policy $\Pi_{\text{Teleop}}$ trained on the teleoperation data. 

\begin{figure}
    \centering
    \includegraphics[width=0.9\linewidth]{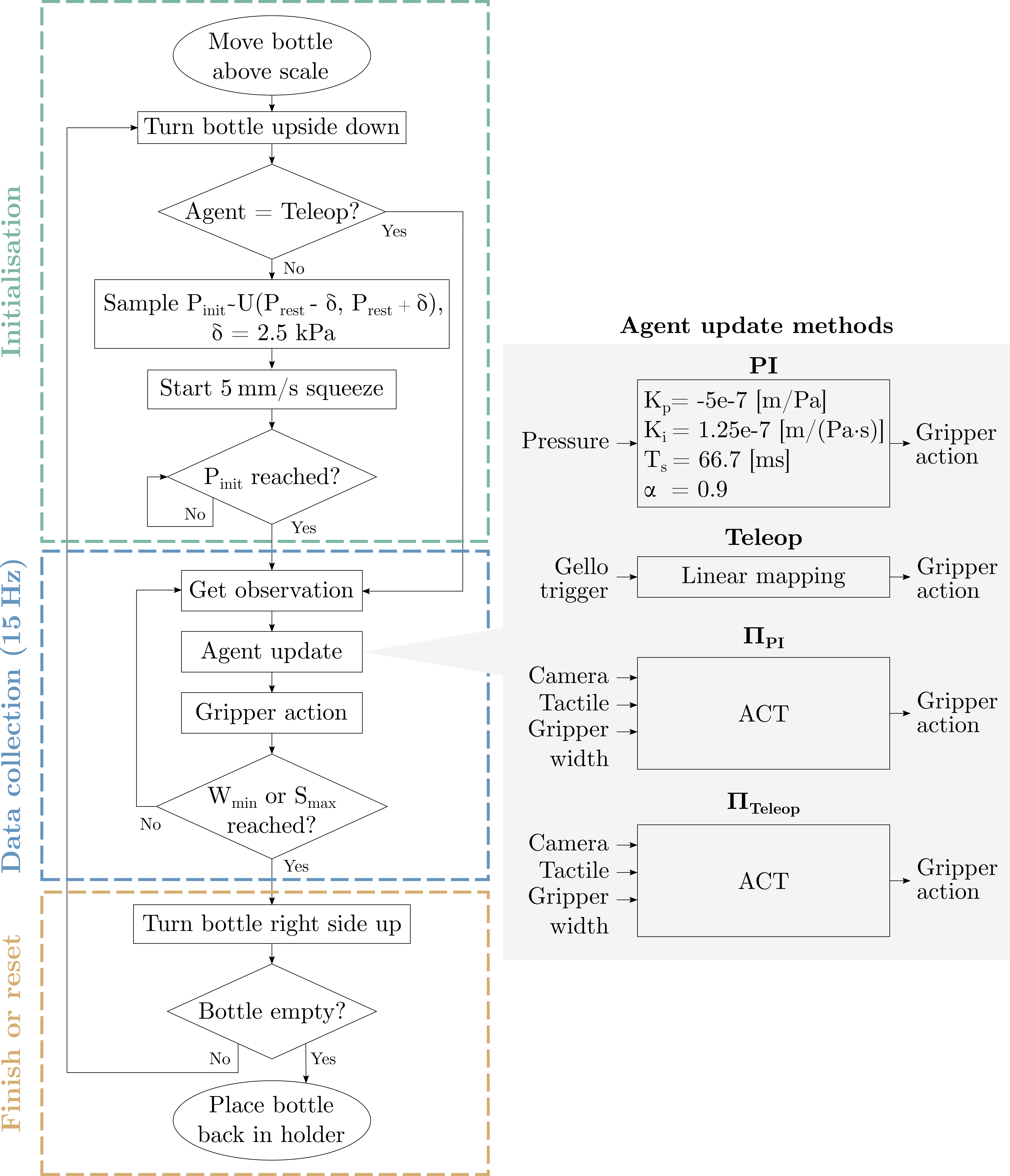}
    \caption{Control flow diagram for data collection.}
    \label{fig:control_flow}
\end{figure}

For these four datasets, the data collection procedure is illustrated in Fig.~\ref{fig:control_flow}.
First, the robot grasps the bottle from the holder and moves it above the Kern scale (see Fig.~\ref{fig:setup1}).
Then, an initial pressure P$_\text{init}$ is determined at random. 
P$_\text{init}$ is to be reached before the agent attempts to maintain a constant flow, such that the training dataset contains trials with a variety of initial conditions (zero flow, drizzle, high flow).
For the PI training set and the evaluation sets of $\Pi_{\text{PI}}$ and $\Pi_{\text{Teleop}}$, P$_\text{init}$ is sampled from a uniform distribution with a 5\,kPa window around the pressure reading when the bottle is not squeezed, P$_\text{rest}$.
With P$_\text{init}$ determined, the gripper starts squeezing at 5\,mm/s until P$_\text{init}$ is reached, at which point the agent attempts to maintain a constant flow.
The human teleoperator, not having access to instrumentation data, was tasked to start half of the trials from zero flow, half with a flow larger than what they would eventually settle on. 
%In the latter case, the teleoperation data was processed to exclude the initial squeeze, as it belongs to task initialisation, not execution.
All teleoperation trials are collected by the same person.

After initialisation, a 15\,Hz control loop starts.
Every iteration, an observation is recorded, composed of an RGB frame of the wrist camera, the internal pressure in the bottle, the current gripper opening, the tactile forces, and the Gello trigger position (teleoperation only).
In addition, the weight reading of the Kern scale is recorded at 3\nobreakdash-4\,Hz.
The scale reading is only used as an extra evaluation of the liquid flow stability, its readout frequency is too low for use as the PI controller's process variable.
Based on this observation, the agent will compute the required gripper action.
The agent update method differs among the four agents (PI controller, human teleoperator, $\Pi_{\text{PI}}$, $\Pi_{\text{Teleop}}$):

\begin{itemize}[leftmargin=*]
     
\item\textbf{PI:} The relative gripper movement $u[k]$ in mm at timestep $k$ is computed as:
    \begin{equation}
    \begin{cases}
        u[k]=\text{K}_\text{p}e[k]+I[k] \\
        I[k]=\alpha I[k-1]+\text{K}_\text{i}\text{T}_\text{s}e[k]
    \end{cases}
    \end{equation}
    K$_\text{p}$ and K$_\text{i}$ are the proportional and integral gain respectively. $e[k]$ is the error on the process variable, which is the pressure in the instrumented bottle. The setpoint is 1.2\,kPa above P$_\text{rest}$, which resulted in the lowest flow while still maintaining a constant stream.
    Specifying the setpoint with respect to P$_\text{rest}$ ensures that changes in air pressure due to weather conditions have no effect.
    The sample time T$_\text{s}$ is the inverse of the control frequency (15\,Hz), and $\alpha$ is an exponential back-off factor to reduce the impact of previous errors.
    The values used for these parameters are included in Fig.~\ref{fig:control_flow}.
    
\item\textbf{Teleop}: 
    The human operator looks at the bottle and attempts to keep the stream constant, squeezing the Gello trigger accordingly.
    The encoder in the Gello trigger outputs an angle that is linearly mapped to the Schunk opening width, such that a released trigger corresponds to a loose grasp of the bottle, and a fully squeezed trigger corresponds to the minimum gripper opening W$_\text{min}$ of 2\,cm.
    The operator chose the smallest flow that still maintained a constant stream, mimicking the chosen setpoint for the PI controller.
    
\item\textbf{$\Pi_{\text{PI}}$ \& $\Pi_{\text{Teleop}}$}: These are neural networks (NN), see section~\ref{ss:learningarch}. An agent update constitutes a forward pass of the NN.
\end{itemize}

A squeezing trial lasts until the maximum scale weight S$_\text{max}$ of 25\,g or the minimum gripper opening W$_\text{min}$ has been reached.
The bottle is turned right side up to equalise the air pressure inside.
If the bottle is empty, it is placed down to be manually refilled.
If it is not yet empty, the robot turns the bottle upside down again, and a new trial starts immediately.
Detecting whether the bottle is empty can be done by comparing the pressure readings before and after turning the bottle over: remaining liquid will add to the pressure reading when the bottle is upside down.
If both readings are approximately equal, the bottle is empty.

For each of the training datasets collected with the PI and Teleop agents, 42 trials were collected, which corresponds to five full bottles squeezed until empty.
For the evaluation datasets, collected with the $\Pi_{\text{PI}}$ and $\Pi_{\text{Teleop}}$ agents, four full bottles were squeezed until empty, resulting in 32 trials each.

\subsection{Learning Architecture} \label{ss:learningarch}
$\Pi_{\text{PI}}$ and $\Pi_{\text{Teleop}}$ are Action Chunking Transformer (ACT) models~\cite{act} with a ResNet18 vision backbone.
They both predict a relative gripper move, given an observation that consists of a 340x480 wrist camera image and a 13-dimensional state vector. 
The state vector is composed of the current gripper opening and the 3D force readings of the four taxels in the tactile fingertip.
We set the chunk size to 10, such that the model predicts a sequence of 10 actions, which corresponds to 0.6 seconds of wall-clock time at 15\,Hz.
Training a model took about 1.5\,h on an NVIDIA RTX4080 GPU. 
Inference takes 12\,ms on an NVIDIA 3090 GPU. %, which is fast enough to achieve our control frequency of 15\,Hz.
% action execution details and differences with ACT
Unlike the original ACT paper, we do not use temporal ensembling nor do we execute multiple actions for each chunk: we found that this highly dynamic task benefited from frequent predictions, and only execute the first action of every predicted chunk. % Predicting a chunk of 10 actions still allows the model to find longer-term correlations between observations and actions during training, and to come up with consistent and effective action sequences during inference, even though only the first action is executed.

\subsection{Evaluation}

The measure of quality for a squeeze trial is the standard deviation $\sigma$ of the time series pressure reading as a proxy for the flow rate.
The first two seconds of a trial are discarded, to exclude transient effects.
We refer to $\sigma$ as the \enquote{score}.

\section{Results \& Discussion}

\subsection{Task Characterisation}

\begin{figure}[tpb]
\centering
%\vspace{2.7mm}
\begin{subfigure}[t]{0.48\textwidth}
    \centering
    \includegraphics[width=\linewidth]{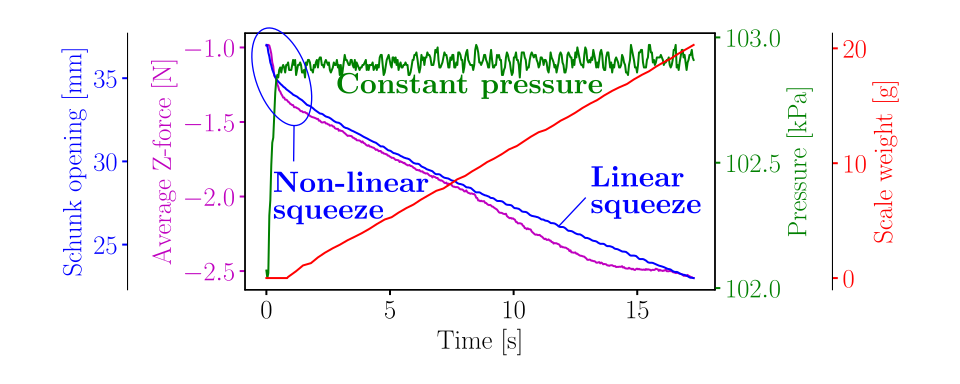}
    \caption{The PI controller settles on a gripper speed of 0.6\,mm/s.}
    \label{fig:0p608108mmps_measured}
\end{subfigure}
\begin{subfigure}[t]{0.48\textwidth}
    \centering
    \includegraphics[width=\linewidth]{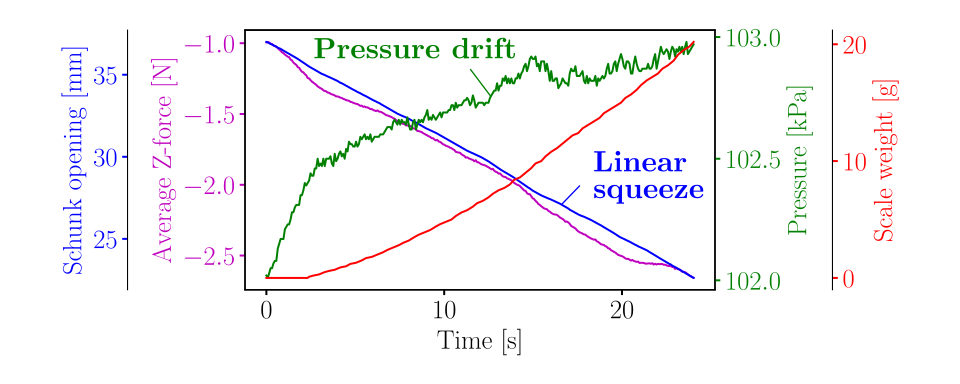} 
    \caption{Forcing a gripper speed of 0.6\,mm/s from the start results in an internal pressure drift.}
    \label{fig:0p608108mmps_forced}
\end{subfigure}
\caption{A constant flow requires non-linear squeezing.}
\label{fig:nonlinear}
\end{figure}

Fig.~\ref{fig:0p608108mmps_measured} shows a typical trial of the PI controller.
This trial started from a P$_\text{init}$ below the setpoint, so the gripper quickly closes at the start, increasing the internal pressure.
The PI controller then maintains an approximately constant pressure, with a $\sigma$ of 26\,Pa. 
The scale clearly shows that the liquid flow is constant: the weight curve rises at 1.22\,g/s with a root mean square error (RMSE) of 0.07\,g.
It appears that the PI controller settles on a linear squeeze of 0.6\,mm/s.
This raises the question whether it is sufficient to squeeze at a constant speed, rather than attempting to learn reactive behaviour.
Fig.~\ref{fig:0p608108mmps_forced} shows that reactive behaviour is necessary: if the gripper squeezes at a constant linear speed of 0.6\,mm/s from the start, the internal pressure drifts.
The RMSE of a linear fit on the scale weight curve increases more than tenfold, to 0.85\,g.
A similar result was obtained for speeds of 1.1 and 1.25\,mm/s.
The gripper opening in Fig.~\ref{fig:0p608108mmps_measured} resembles the manually tuned finger motion used in~\cite{huang2024}.

%The learned agents $\Pi_{\text{PI}}$ and $\Pi_{\text{Teleop}}$ have indeed adopted this reactiveness: Fig.~\ref{fig:reactiveness} shows two $\Pi_{\text{PI}}$ trials, one starting from zero flow (Fig.~\ref{fig:trial_from_low}-\ref{fig:wrist_img_trial_from_low_obs75}), another from a flow stronger than the setpoint (Fig.~\ref{fig:trial_from_high}-\ref{fig:wrist_img_trial_from_high_obs75}).
%The learned policy recognises the flow rate and adjusts as needed, instead of merely paying attention to proprioception and maintaining a constant gripper speed.

%\begin{table}[]
%\centering
%\caption{Dataset sizes. }%The laser and microphone perform similar for the ambient sound experiments but the laser outperforms the microphone significantly on the targeted disturbance experiments.}
%\label{tab:dataset_sizes}

%\setlength{\tabcolsep}{2pt}
% Using booktabs for better spacing and visual separation
%\begin{tabular}{lc}
%\toprule % Use \hline if not using booktabs
%\textbf{Agent}  & \textbf{Number of trials} \\ 
%\midrule
%Manual & 34 \\
%PI & 42 \\
%Teleop & 42 \\
%$\Pi_{\text{PI}}$ & 32 \\
%$\Pi_{\text{Teleop}}$ & 32 \\
%\bottomrule % Use \hline if not using booktabs
%\end{tabular}
%\end{table}

%\begin{figure}[tpb]
%\centering
%\vspace{2.7mm}
%\begin{subfigure}[t]{0.28\textwidth}
%    \centering
%    \includegraphics[width=\linewidth]{img/trial_from_low.png}
%    \caption{A trial starting from zero initial flow.}
%    \label{fig:trial_from_low}
%\end{subfigure}\hfill
%\begin{subfigure}[t]{0.085\textwidth}
%    \centering
%    \includegraphics[width=\linewidth]{img/wrist_img_trial_from_low_obs0.png}
%    \caption{View at 0s in (a).}
%    \label{fig:wrist_img_trial_from_low_obs0}
%\end{subfigure}\hfill
%\begin{subfigure}[t]{0.085\textwidth}
%    \centering
%    \includegraphics[width=\linewidth]{img/wrist_img_trial_from_low_obs75.png}
%    \caption{View at 5s in (a).}
%    \label{fig:wrist_img_trial_from_low_obs75}
%\end{subfigure}
%\vspace{2mm}
%\begin{subfigure}[t]{0.28\textwidth}
%    \centering
%    \includegraphics[width=\linewidth]{img/trial_from_high.png}
%    \caption{A trial starting from high flow.}
%    \label{fig:trial_from_high}
%\end{subfigure}\hfill
%\begin{subfigure}[t]{0.085\textwidth}
%    \centering
%    \includegraphics[width=\linewidth]{img/wrist_img_trial_from_high_obs0.png}
%    \caption{View at 0s in (d).}
%   \label{fig:wrist_img_trial_from_high_obs0}
%\end{subfigure}\hfill
%\begin{subfigure}[t]{0.085\textwidth}
%    \centering
%    \includegraphics[width=\linewidth]{img/wrist_img_trial_from_high_obs75.png}
%    \caption{View at 5s in (d).}
%    \label{fig:wrist_img_trial_from_high_obs75}
%\end{subfigure}
%\caption{The learned policies recover from too low as well as too high flows.}
%\label{fig:reactiveness}
%\end{figure}

\subsection{Agent Comparison}

\begin{figure}
\centering
\includegraphics[width=\linewidth]{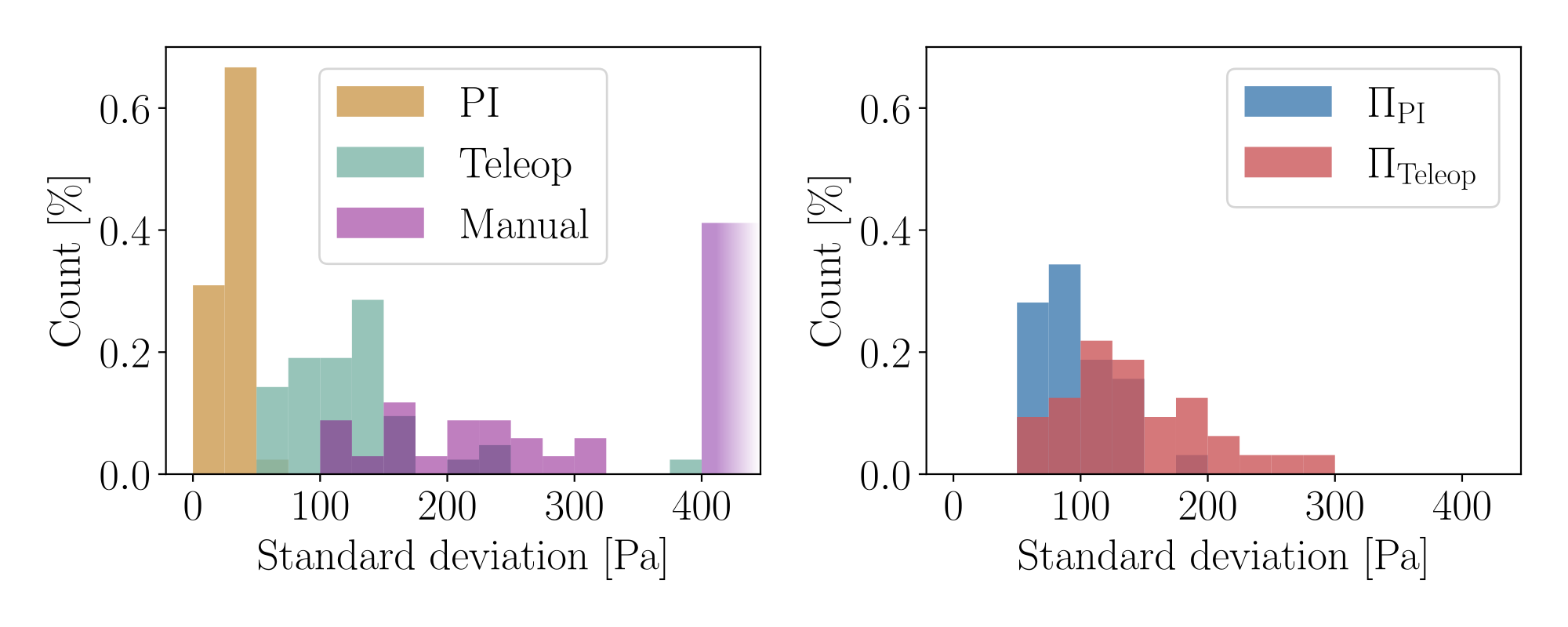}
\caption{Comparing agent scores.}
\label{fig:histograms}
\end{figure}

For each collected trial, the standard deviation $\sigma$ is computed.
The results are compiled in histograms with a bin width of 25\,Pa, see Fig.~\ref{fig:histograms}.
The manual trials show a very wide spread, some even reaching a $\sigma$ of over 1000. 
This is because some people naturally performed better than others, but also because people tried different squeezing strategies, some of which were ineffective.
For clarity of the figure, the final histogram bin aggregates all scores over 400.
The lowest $\sigma$ achieved was 105, which serves as a reference point to compare the performance of other agents.
Notably, 40\,\% of the teleoperation demonstrations score better than the best-case manual performance. 
This is attributed to the Gello trigger taking less force to squeeze than the bottle itself, making it physically less demanding for the teleoperator.
On average, the teleoperated trials had a $\sigma$ of 126$\pm$58\,Pa.
We argue that our teleoperated demonstrations are \enquote{good}, in the sense that they are comparable to best-case manual performance.
The PI training set scores by far the best, with an average $\sigma$ of 30$\pm$7\,Pa.

The evaluation set of the $\Pi_{\text{PI}}$ policy has a mean score of 93$\pm$30\,Pa, compared to 144$\pm$56\,Pa for the $\Pi_{\text{Teleop}}$ policy.
Assuming Gaussian priors, we can calculate the probability that $\Pi_{\text{PI}}$ performs better than $\Pi_{\text{Teleop}}$.
Let $X_\text{PI}$ be the $\sigma$ of a random $\Pi_{\text{PI}}$ trial and $X_\text{Teleop}$ of a $\Pi_{\text{Teleop}}$ trial:
%\begin{equation}
%\begin{cases}
%    X_\text{PI}\sim N(93, 30) \\
%    X_\text{Teleop}\sim N(144, 56)
%\end{cases}
%\end{equation}
%\begin{equation} \label{e:combined_gaussian}
%    \Rightarrow X_\text{Teleop}-X_\text{PI}\ \sim N(51, 64)
%\end{equation}
\begin{equation}\label{e:combined_gaussian}
    \left\{
    \begin{array}{l}
    X_\text{PI} \sim N(93, 30) \\
    X_\text{Teleop} \sim N(144, 56)
    \end{array}
    \right.
    \Rightarrow
    X_\text{Teleop} - X_\text{PI} \sim N(51, 64)
\end{equation}
\begin{equation}
\begin{split}
    \text{P}[\Pi_{\text{PI}}~\text{better than}~\Pi_{\text{Teleop}}] 
    & = \text{P}[X_\text{PI}<X_\text{Teleop}] \\
    & = \text{P}[X_\text{Teleop} - X_\text{PI} > 0] \\
    &  \stackrel{(\ref{e:combined_gaussian})}{=} 78\,\%
\end{split}
\end{equation}
We conclude that, given the choice between $\Pi_{\text{PI}}$ and $\Pi_{\text{Teleop}}$, choosing $\Pi_{\text{PI}}$ will lead to a better result in 78\,\% of cases.
This is an added benefit on top of the partial automation of demonstrations for $\Pi_{\text{PI}}$, because of which data collection required significantly less human effort.
However, instrumentation requires hardware design effort before data collection can start.
This raises important trade-offs: Is the performance gain achieved by exploiting instrumentation worth the initial design effort?
Is data collection prohibitively time consuming at scale, so that an initial investment in instrumentation can save time overall?
The answer to these questions may change on a case-by-case basis, but we believe that, in the search for generalist agents, instrumentation has a lot of unexplored potential.

\section{Conclusion and Future Work}
We developed an instrumented squeeze bottle to aid in learning the task of squeezing a constant flow from the bottle. 
Five datasets were collected.
It was found that manual human performance varied wildly, and that human teleoperation can be easier than manual operation if the required grasping force to actuate the teleoperation mechanism is less than for the bottle itself.
Still, a PI controller exploiting the instrumentation data significantly outperformed the teleoperator.
In addition, like~\cite{maram2022} and \cite{bilal2024}, we explicitly showed that better demonstrations lead to a better policy: an ACT policy trained on the automated PI demonstrations performs better than an ACT policy trained on the teleoperated demonstrations in 78\,\% of cases.
Hence, instrumentation not only reduces human effort during data collection, but can also improve the performance of learned policies.

In future work, the generalisation capabilities of the policies should be evaluated.
Our results are promising, but for practical applications the true quality of our squeezing policy should be measured on out-of-distribution bottles and environments.
%Second, we did not consider the relative contribution of each of the model inputs (vision, tactile, and proprioception).
%Such results can be interesting for assessing the relevance of tactile sensing, 
In addition, we have shown just one possible approach for incorporating instrumentation in imitation learning.
Our aim is to develop similar case studies to build an overview of the possible benefits of instrumentation for robot learning.

\balance
\bibliographystyle{IEEEtran}
\bibliography{references.bib}

\end{document}